\documentclass[10pt,twocolumn,letterpaper]{article}

\usepackage{wacv}      

\usepackage{graphicx}
\usepackage{amsmath}
\usepackage{amssymb}
\usepackage{booktabs}

\usepackage[table]{xcolor}
\usepackage[accsupp]{axessibility} 
\usepackage{footmisc}
\usepackage{tikz}

\usepackage{newfloat}
\usepackage{listings}

\usepackage[pagebackref,breaklinks,colorlinks]{hyperref}

\usepackage[capitalize]{cleveref}
\crefname{section}{Sec.}{Secs.}
\Crefname{section}{Section}{Sections}
\Crefname{table}{Table}{Tables}
\crefname{table}{Tab.}{Tabs.}

\usepackage{pifont}
\usepackage{multirow}
\usepackage{lipsum}

\newcommand*\rot{\rotatebox{90}}
\newcommand*\OK{\ding{51}}
\def\wacvPaperID{2398} 
\def\confName{WACV}
\def\confYear{2024}

\begin{document}

\title{IDD-AW: A Benchmark for Safe and Robust Segmentation of Drive Scenes\\
in Unstructured Traffic and Adverse Weather}


\author{Furqan Ahmed Shaik$^{*1}$ \qquad Abhishek Reddy Malreddy$^1$ \qquad Nikhil Reddy Billa$^1$ \\
Kunal Chaudhary$^2$ \qquad Sunny Manchanda$^2$ \qquad Girish Varma$^1$\\[0.5em]
$1.$ Machine Learning Lab, IIIT Hyderabad \qquad $2.$ DYSL-AI, DRDO\\[0.5em]
{
*furqan.shaik@research.iiit.ac.in}
}

\maketitle

\begin{abstract}
Large-scale deployment of fully autonomous vehicles requires a very high degree of robustness to unstructured traffic, weather conditions, and should prevent unsafe mispredictions. While there are several datasets and benchmarks focusing on segmentation for drive scenes, they are not specifically focused on safety and robustness issues.
We introduce the IDD-AW dataset, which provides 5000 pairs of high-quality images with pixel-level annotations, captured under rain, fog, low light, and snow in unstructured driving conditions. As compared to other adverse weather datasets,
we provide i.) more annotated images, ii.) paired Near-Infrared (NIR) image for each frame, iii.) larger label set with a 4-level label hierarchy to capture unstructured traffic conditions. We benchmark state-of-the-art models for semantic
segmentation in IDD-AW. We also propose a new metric called “Safe mean Intersection over Union (Safe mIoU)” for hierarchical datasets which penalizes dangerous mispredictions that are not captured in the traditional definition of mean Intersection over Union (mIoU). The results show that IDD-AW is one of the most challenging datasets to date for these tasks. The dataset and code will be available here: \url{http://iddaw.github.io}.

\end{abstract}
\section{Introduction}
\label{sec:intro}
\begin{figure}[t]
\centering
\includegraphics[width=\columnwidth]
{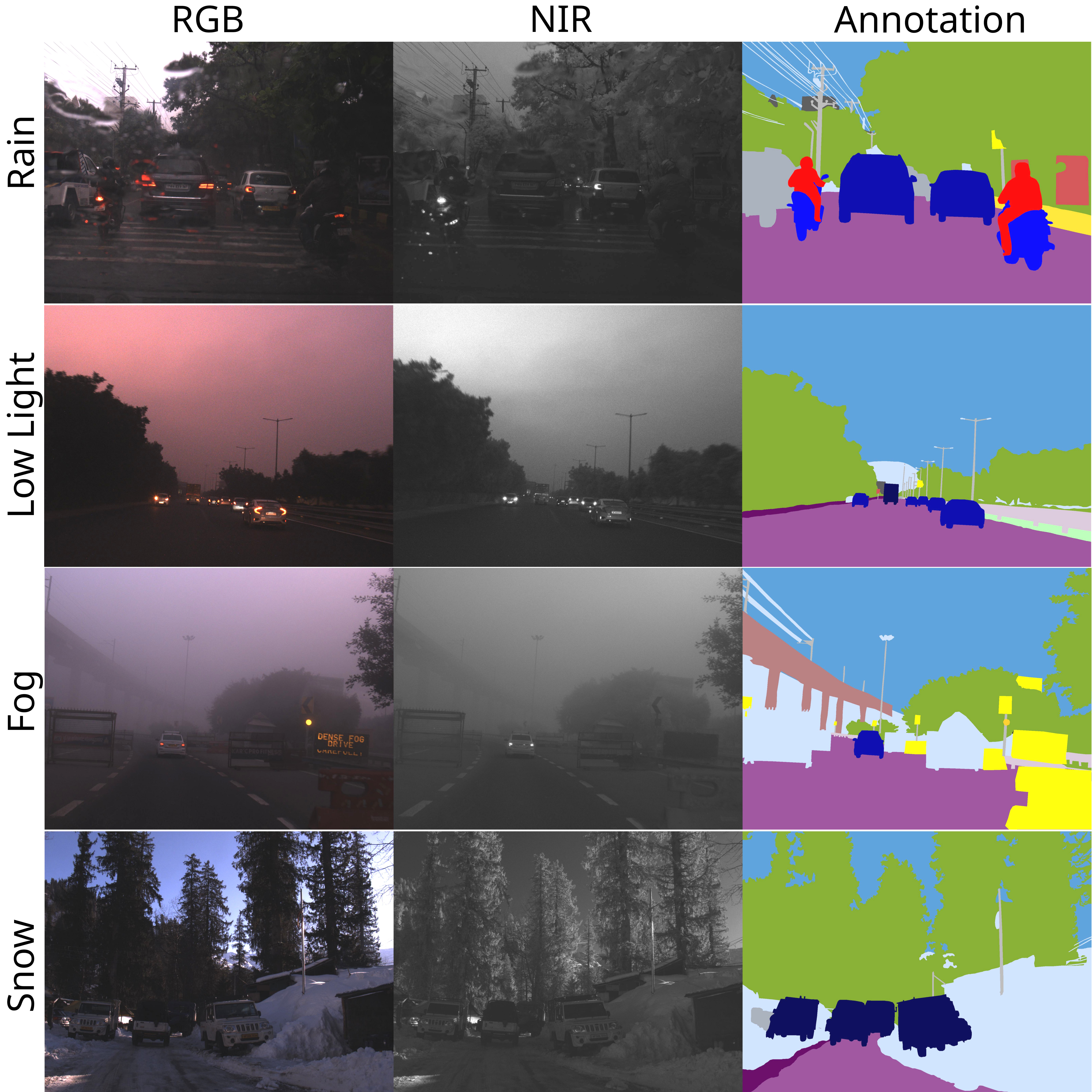} 

\caption{Examples of RGB/NIR/Annotations triplets in four adverse weather conditions from IDD-AW. The color legend for annotation is given in Figure \ref{fig:idd_ds_pixel_comparison}.}

\label{fig:ds-front}
\end{figure}

    Semantic Segmentation of driving scenes is one of the major tasks in the development of autonomous driving vehicles. Even though there has been much work in terms of datasets \cite{cordts2015cityscapes,cordts2016cityscapes,kitti,mapillary}, there are still many challenges when it comes to the representation of rare events and conditions. One of the major challenges is the robustness of the model to high amounts of traffic participants and adverse weather conditions. While state-of-the-art models give high prediction accuracies, it's unclear how they will perform in such cases. Vehicles with full autonomy cannot be deployed safely unless these corner cases are analyzed.  

    Secondly, given the focus on safety in autonomous vehicles, it is essential to check if the metrics used accurately represent the safety concerns in road scenes. Traditionally segmentation models have been evaluated based on the mIoU which is a generic metric for the quality of the segmentation. However, mIoU doesn't capture safety concerns specific to road scenes. For example, accurate prediction of traffic participants  (vehicles, living things) and roadside objects are more essential compared to far objects. Also for autonomous driving, certain types of mispredictions are more tolerable than others. For example, a car being mispredicted as a bus is more tolerable than a car being mispredicted as a road. mIoU treats both these mispredictions in the same way while they represent different severity of safe mispredictions.

    In this paper, we introduce the IDD-AW  dataset which captures highly unstructured traffic in adverse weather conditions. Here, highly unstructured traffic refers to situations where road conditions are intricate and challenging to predict, often characterized by a diverse mix of vehicles, pedestrians, and other dynamic elements. The dataset captures driving scenarios under various adverse weather conditions, encompassing rain, fog, low light, and snow. The distribution of pixels and instances of traffic participants is higher when compared to existing datasets. Additionally, we provide paired Near Infrared (NIR) images for every frame, thereby increasing the information available to models and enhancing their ability to make precise predictions.

    We also propose a new metric called "Safe mIoU (SmIoU)" which penalizes dangerous mispredictions that are not captured in the traditional mIoU metric. We propose this metric for all hierarchical datasets where the severity of the penalty is based on the tree distance. We also calculate this metric over various label sets varying from the complete set and narrowing it down to several important classes such as traffic participants and roadside objects which directly impact autonomous driving. 

    We benchmark models on the IDD-AW  dataset and compare them with results on other similar datasets like Cityscapes\cite{cordts2016cityscapes}, IDD \cite{idd}, ACDC \cite{acdc}. We show that appending the NIR image to the input increases the accuracy of our dataset. We also show high contrast in results when comparing mIoU and SmIoU using the same trained models. We provide qualitative and logical explanations, that the lowering of Safe mIoU as compared to mIoU exposes safety concerns in existing models.

    \paragraph{Main Contributions.}
    \begin{enumerate}
        \item We propose an adverse weather conditions dataset for robust segmentation in unstructured traffic scenes (see Section \ref{sec:dataset}). 
        
        \item We also propose a new metric called "Safe mIoU" to ensure the penalizing of misclassified pixels in hierarchical datasets (see Section \ref{sec:safemiou}).
        
        \item We show both qualitative and quantitative results to prove our dataset is more diverse and challenging than other existing datasets (see Section \ref{sec:results}). 
    
    \end{enumerate}

\section{Related Work}

\begin{table}[t]
\centering
\resizebox{\columnwidth}{!}{
\begin{tabular}{@{}lrrrrrrrrr@{}}
\toprule
    Dataset  & \rot{\begin{tabular}[c]{@{}l@{}}Labeled\\ images\end{tabular} } &\rot{Rain} &\rot{Fog} & \rot{Snow} & \rot{Lowlight} & \rot{Labels} & \rot{NIR}\\
\midrule

     \begin{tabular}[c]{@{}l@{}}Foggy\\ Driving\end{tabular} & 101 & 0 & 101 & 0 & 0 & 19 & ~ \\ \midrule
    \begin{tabular}[c]{@{}l@{}}Foggy\\ Zurich\end{tabular} & 40 & 0 & 40 & 0 & 0 & 19 & ~ \\ \midrule
    \begin{tabular}[c]{@{}l@{}}Nighttime\\ Driving\end{tabular} & 50 & 0 & 0 & 0 & 50 & 19 & ~ \\ \midrule
    \begin{tabular}[c]{@{}l@{}}Dark\\ Zurich\end{tabular} & 201 & 0 & 0 & 0 & 201 & 19 & ~ \\ \midrule
    Raincouver & 326 & 326 & 0 & 0 & 95 & 19 & ~ \\ \midrule
    WildDash & 226 & 13 & 10 & 26 & 13 & 19 & ~ \\ \midrule
    BDD100K & 1346 & 213 & 23 & 345 & 765 & 19 & ~  \\ \midrule
    ACDC  &  4006 & 1000 & 1000 & 1000 & 1006 & 19 & ~  \\ \midrule

    \textbf{IDD-AW}    & \textbf{5000} & \textbf{1500} & \textbf{1500} & \textbf{1000} & \textbf{1000} & \textbf{30} & \OK \\ 
    \bottomrule

  \end{tabular}
}
\caption{Comparison of IDD-AW  with other adverse weather condition datasets. IDD-AW  provides i.) more labeled images ii.) with more (30) labels that capture unstructured driving conditions and iii.) NIR images. }
\label{tab:overall-datasets-comp}
\end{table}

\paragraph{Road scene segmentation datasets.}
The Cityscapes Dataset{\cite{cordts2016cityscapes, cordts2015cityscapes}} is one of the most popular datasets for driving scene segmentation. Along with it, there are other driving datasets such as KITTI \cite{kitti}, Pascal VOC{\cite{pascalvoc}, Mapillary vistas{\cite{mapillary}} etc which contain a large collection of images. However, most of these datasets are captured in structured environments and have label sets that cannot completely satisfy the unstructured scenarios. IDD{\cite{idd}} introduced a large-scale annotated dataset in unstructured driving conditions with more than 10K annotated images. It contains drive scenes with a larger label set when compared to Cityscapes and includes 2-wheelers, pedestrians, and roadside objects. 
However, none of the above datasets have images in adverse weather conditions. All the datasets have been captured in clear daylight conditions with minimum image distortion. 

\begin{figure*}[t]
\centering
\includegraphics[width=\textwidth]
{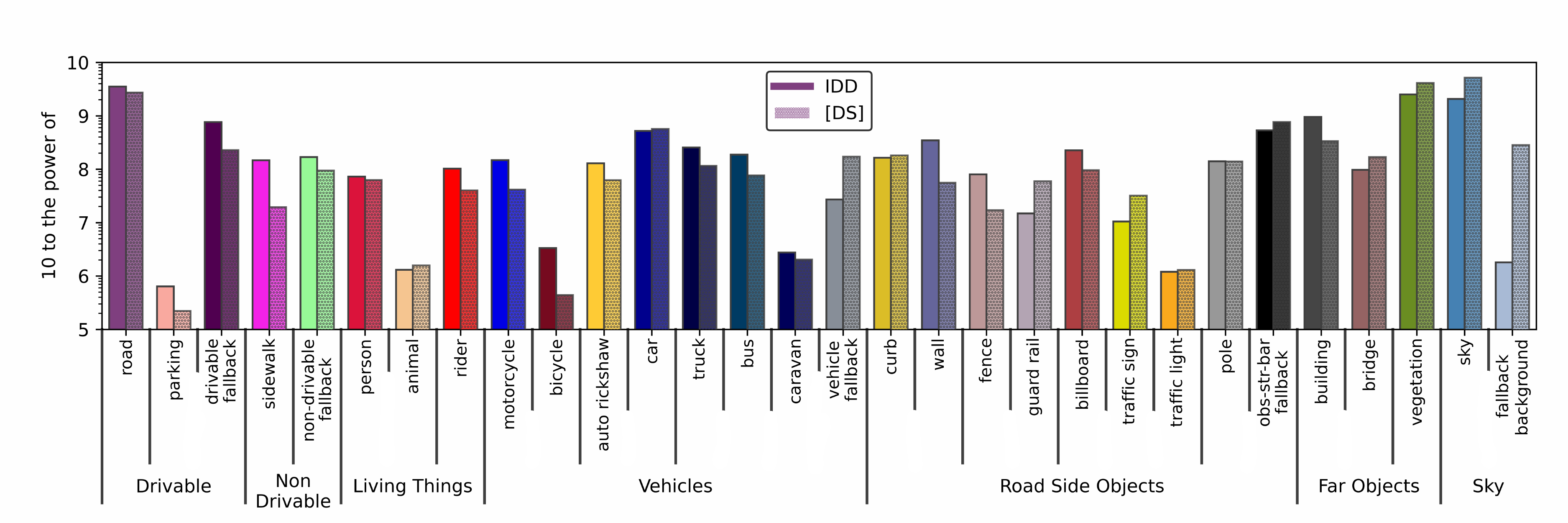} 
\caption{IDD \& IDD-AW  hierarchical labels (x-axis) and pixel count (y-axis) comparison. IDD provides data in unstructured driving conditions and our pixel composition is similar to it.  Both datasets have 30 labels and a 4-level label hierarchy.  The colors used will serve as legends for all plots of annotations and predictions in this paper. Normalization is done relative to image resolution and number of images in the datasets.}
\label{fig:idd_ds_pixel_comparison}
\end{figure*}

\paragraph{Adverse weather road scene datasets.}
Oxford Robotcar \cite{oxford_robotcar} was the first
real-world large-scale dataset in which adverse visual conditions such as nighttime, rain, and snow were significantly
represented, but it did not feature semantic annotations. Foggy Zurich{\cite{foggy_zurich}}, Foggy Cityscapes{\cite{foggy_driving}} have synthesized fog images generated by applying a mask over the original images. Rain1400\cite{rain1400}, RainyCityscapes \cite{rainy_cityscapes} Rain100H\cite{rain100h}, and Rain12\cite{rain12} are prime examples for synthetic rain datasets. Comprehensive Snow Dataset \cite{snow_csd} is another example of synthetic snow images. 
Datasets such as Rainy Wcity \cite{rainy_wcity} BDD100K{\cite{bdd100k}}, ACDC{\cite{acdc}}, WildDash \cite{wilddash}, WildDash2 \cite{Wilddash2} and Ithaca365{\cite{Ithaca_365}} have real images which have been captured in various adverse conditions like rain, fog, snow and low light/night time. However, none of these datasets have both NIR and RGB components which could help us give a better insight into the images in these adverse conditions. We compare these datasets with IDD-AW  in Table \ref{tab:overall-datasets-comp}.

\paragraph{Datasets with NIR/Multispectral Data.}
Nuscenes \cite{caesar2020nuscenes}, the first dataset to carry the fully autonomous vehicle sensor suite: 6 cameras, 5 radars, and 1 lidar, all with a full 360-degree field of view. The Multi-Spectral Road Scenarios (MSRS) dataset \cite{piafusion} focuses on road-related scenarios by combining NIR and RGB modalities. The TNO Hyperspectral Dataset \cite{tno_dataset} provides NIR and RGB images obtained from an airborne platform. Similarly, the KAIST Multispectral Pedestrian Dataset \cite{kaist_dataset} offers synchronized NIR and RGB images captured in urban settings. IDD-3D \cite{idd_3d}, which consists of multimodal data from multiple cameras and LiDAR sensors across various traffic scenarios. However, most of these datasets do not cover adverse weather images. IDD-AW dataset provides synchronized RGB and NIR images in adverse weather and unstructured driving scenarios. 


\paragraph{Safe \& Robust Segmentation.}
The robustness of a dataset can also be viewed in terms of the ambiguity of labels in the dataset and the misclassification of corner cases. Disambiguity of labels can be achieved by introducing a hierarchy in the dataset labels. Hierarchical semantic segmentation explains visual scenes with multi-level abstraction by considering a structured class label hierarchy. 

Li et al. \cite{li2018hierarchical} addresses hierarchical segmentation at both scene level and object level.} Shaik et al. \cite{shaik2023accelerating} aims to accelerate semantic segmentation and other computer vision tasks on GPUs using sparsity frameworks. In the work of Li et al. \cite{li2022deep}, the authors aim to enhance hierarchical semantic segmentation by modifying the cross-entropy loss and penalizing it over various class hierarchies. 
In the research by Bar et al. \cite{robust_redundant}, the `teacher-student framework' is employed, where a pre-trained model guides the training of another model, a common practice to improve segmentation. 
In the study by Kamann et al. \cite{kamann2021benchmarking}, the authors examine the robustness of semantic segmentation networks concerning a wide range of image corruptions based on their architecture.

IDD-AW contains images in unstructured driving scenarios which are also captured in various adverse weather conditions. This signifies the robustness of the dataset along various conditions. Additionally, we introduce a metric called Safe-mIoU which penalizes unsafe mispredictions more than traditional mIoU. This helps in improving the safety of the models especially for hierarchical driving datasets.

\begin{figure*}[t]
\centering
\includegraphics[width=0.98\textwidth]{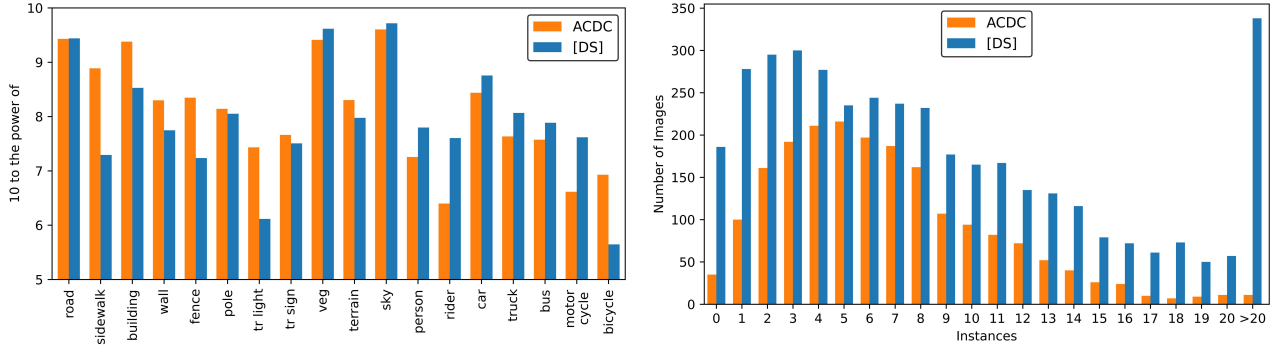}
\caption{Comparison of ACDC \& IDD-AW: Pixel wise (left), traffic participants instances count (right). Pixel counts are normalized by resolution and the number of images in the datasets. IDD-AW  has more pixels (left) as well as more instances per image (right) of traffic participants (TP) which includes all vehicles and living things that represent unstructured traffic. Note that there are over 300 images in IDD-AW  with more than 20 instances of TP while ACDC has only around 10.}
\label{fig:acdc_IDD-AW }
\end{figure*}

\section{The IDD-AW  Dataset}
\label{sec:dataset}

\paragraph{Data Collection or Acquisition.}

The dataset primarily focuses on capturing traffic scenarios in adverse weather conditions such as rain, fog, snow, and low light, with an emphasis on unstructured driving scenarios. Our dataset comprises a total of 202 driving sequences, providing a comprehensive collection of challenging driving scenarios. The dataset is captured at different locations in India comprising urban, rural, and even hilly areas to ensure diversity across various conditions.

For data capture, we used JAI's FS-3200D-10GE camera. This is a 2-CMOS multispectral prism camera that captures images in different spectral bands simultaneously. Specifically, it provides a visible color channel from 400-670 nm and a near-infrared (NIR) channel from 740-1000 nm. To ensure a clear separation between different adverse conditions, each capture was made under only one adverse scenario to make sure the images were diverse between various conditions.

\paragraph{Frame Selection.}
The frames captured by the camera are filtered using a 3-second threshold to eliminate any repetitive frames. The goal of this threshold is to ensure that unique frames are used for subsequent processing. After the filtering step, the RGB-NIR pairs are matched based on their time stamps. However, to ensure accuracy, the matches are also manually verified by reviewing all frames. From the matched pairs, visually appealing frames that feature a diverse range of traffic participants and showcase adverse driving conditions are shortlisted. This process allows for the selection of high-quality frames for the dataset.

\paragraph{Label hierarchy and Annotation.}
The label set used is the same as in IDD \cite{idd}. This label set has a hierarchical structure with four levels, consisting of 7 labels at level 1 and 30 labels at level 4 (see x-axis of Figure \ref{fig:idd_ds_pixel_comparison}). This adds a higher level of complexity to our dataset when compared to existing datasets like Cityscapes, and even when compared to adverse weather datasets like ACDC or Foggy Cityscapes.
For labelling the dataset, we had a team of highly skilled annotators.
The annotation process for each image typically requires 1.5 to 2.5 hours to complete, encompassing the initial annotation and quality check. To ensure the correctness of the annotations, multiple steps are taken, such as verifying annotations against predefined criteria and conducting a final quality check by experienced annotators.

\paragraph{Dataset Splits.}
IDD-AW  is split into four sets corresponding to the examined conditions. We manually selected 1500 rainy, 1500 foggy, and 1000
lowlight and 1000 snowy images from the
recordings for dense pixel-level semantic annotation, for a
total of 5000 adverse-condition images. The selection process aimed at maximizing the complexity and diversity of captured scenes. 

We used a thorough train-test split technique with set limits to permit robust evaluations. All images from a drive sequence are either completely in train or in test to ensure that the test is truly unseen and distinct. For each weather condition, the test set includes drive sequences in a ratio of 0.18 to 0.22 of total sequences, ensuring adequate representation of varied driving scenarios. The average number of frames per drive sequence in the test set was kept within the ratio of 0.9 to 1.2 of the average frames per drive sequence of the entire dataset. Class-specific fairness is maintained by ensuring that the average instances per image in the test set vary between 0.8 and 1.2 of the overall dataset.
For precise pixel-level accuracy evaluation, the test dataset must meet strict criteria: an average pixel ratio of 0.8 to 1.2 for at least 18 classes, and 0.7 to 1.3 for at least 22 classes.

\paragraph{Comparison with Other Datasets.} In Table \ref{tab:overall-datasets-comp}, we compare IDD-AW to existing datasets that address semantic segmentation under adverse conditions. Most of these datasets are specialized in a single condition and are relatively small in scale. IDD and ACDC are the two primary large-scale datasets, focusing on unstructured driving scenes and adverse weather, respectively. IDD-AW is unique in its focus on drive scenes that are both unstructured and affected by adverse weather. Additionally, we provide a Near-Infrared (NIR) component, recognizing its potential to enhance segmentation accuracy and safety. NIR images tend to exhibit higher contrast and sharpness, while RGB images provide color and luminescence information that enhances the visual appeal. 

Figure \ref{fig:idd_ds_pixel_comparison} illustrates the pixel-wise comparison between IDD and IDD-AW datasets. Although our dataset features images captured in adverse weather conditions, most of the classes exhibit pixel counts similar to those of the IDD dataset which represents unstructured traffic.

Additionally, Figure \ref{fig:acdc_IDD-AW } provides a pixel-wise comparison, as well as traffic participant, counts when compared with ACDC. Our dataset notably contains more pixels for crucial traffic participants such as roads, pedestrians, riders, and cars. Similarly, when comparing instance counts for traffic participants per image, nearly every image in our dataset has more instances than those in ACDC. Moreover, over 300 images in our dataset contain more than 20 instances, while ACDC only includes approximately 10-20 images with the same number of instances. This observation highlights that our dataset features numerous dense images in terms of both pixels per class and traffic participants per image.

\begin{figure}[t]

    \includegraphics[width=\columnwidth]{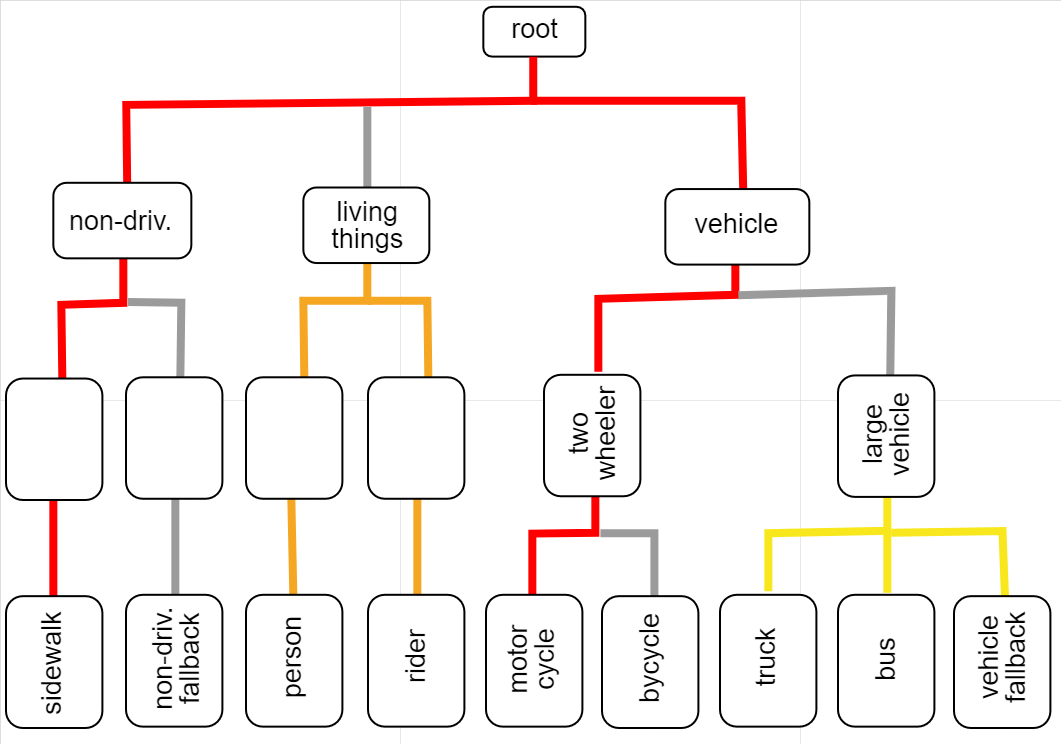}

\caption{Example of tree distances between labels. Tree distance captures the severity of the misprediction using the label hierarchy. The $\text{td}(\text{sidewalk},\text{motorcycle})=3$ since the length of the path is 6 and td is length/2. Similarly $\text{td}(\text{person},\text{rider})=2$ and $\text{td}(\text{truck},\text{bus})=1$.} 
\label{fig:tree}
\end{figure}

\section{Safe-mIoU: A metric for safe segmentation}
\label{sec:safemiou}

\begin{figure}[t]
\centering
\includegraphics[width=\columnwidth]
{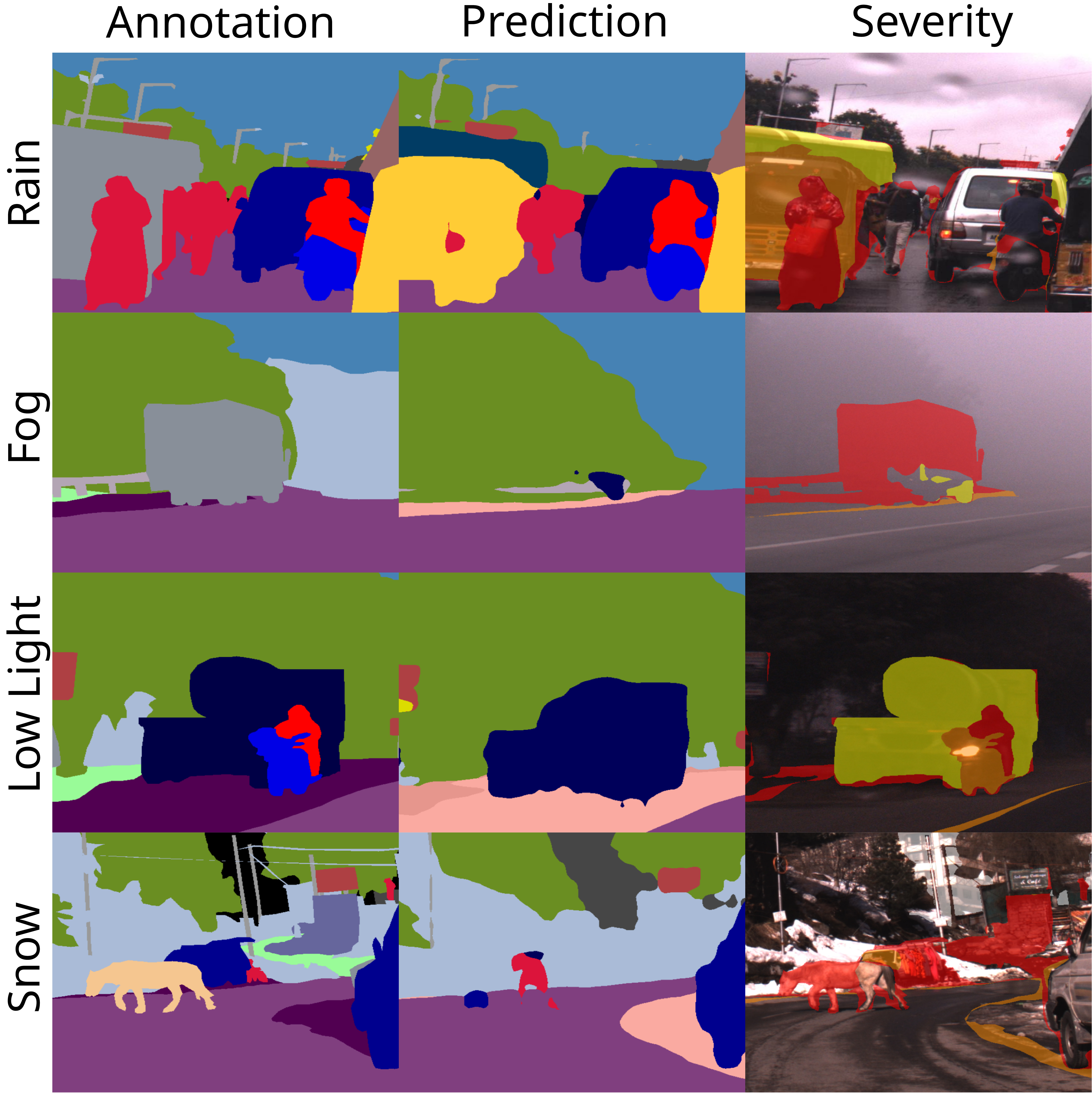} 
\caption{Examples from IDD-AW, showing a pedestrian in the rain, a truck in the fog, a rider/bike in low light, and an animal in the snow are mispredicted respectively. In the Severity Column, Red, orange, and yellow colors denote the level of danger posed by mispredictions. In the third row, a bike being mispredicted as a truck is less dangerous (in orange) than a rider being predicted as a vehicle (in red). mIoU considers these different colored mispredictions in the same way, while the proposed safe mIoU gives a higher penalty for orange and red mispredictions.  In these images, the proposed safe mIoU is less than mIoU by more than 20\%.}

\label{fig:cond_images_display}
\end{figure}

Traditionally, mIoU (mean of Intersection over Union) is a widely used metric for semantic segmentation tasks. It reflects the quality of segmentation by considering both false positives (pixels mistakenly classified as a class) and false negatives (pixels of the class not detected). It captures the balance between precision and recall. However, while mIoU is an excellent general-purpose metric, its limitations become apparent when attempting to quantify the safety of driving scenes. However, mIoU's strength lies in its generality, and this generality becomes its limitation in contexts where class importance as well as the severity of the mispredictions varies. In driving scenes, the safety of pedestrians, vehicles, traffic signs, and other critical elements holds paramount importance. Misclassifying these safety-critical classes can lead to dire consequences. Also, certain misclassifications are dangerous, while others are tolerable. For example, it is tolerable to misclassify one vehicle as another type, while it is dangerous to misclassify a pedestrian or the vehicle as the road (see Figure \ref{fig:cond_images_display}). Unfortunately, traditional mIoU does not inherently capture these safety concerns. It treats all classes as equal entities, disregarding their impact on real-world driving scenarios. 



To address this gap, a more refined metric is needed – one that can not only assess the quality of segmentation results but also consider the severity of unsafe mispredictions. We propose a new \textbf{Safe mIoU (SmIoU)} metric. The calculation of Safe mIoU involves selecting a subset of critical/important classes $C_{\text{imp}}$ from the entire class set $C$. These chosen classes typically encompass those that have a direct impact on safety within driving scenes, such as pedestrians, vehicles, and traffic signs.  Furthermore, we use the distance between the predicted and the actual label in the hierarchical label tree of IDD-AW, to measure the severity of misclassifications in $C_{\text{imp}}$. It aims to provide a more holistic evaluation that aligns with the priorities of driving scene applications and the overall safety of the environment.

The essence of SmIoU lies in the introduction of hierarchical penalty, a strategy that takes into account the semantic relationships between classes. Misclassifications within critical classes, and non-critical classes classified as critical, are penalized based on their distance in the class hierarchy. The \emph{tree distance} (td) between a pair of labels is the length of the shortest path in the class hierarchy tree divided by 2 (see Figure \ref{fig:tree}). So the td between person and rider is 1, while sidewalk and motorcycle is 3 (see the class hierarchy in \ref{fig:idd_ds_pixel_comparison}), denoting that the latter is a more severe misprediction than the former. The calculation of SmIoU involves computing the $I_c^{\text{safe}}$ for each class as shown in equation \ref{eq:1}. The safe IoUs are the IoUs with a penalty for misclassification of that class weighted by the tree distance. The final SmIoU score is obtained by taking the mean of these $I_c^{\text{safe}}$ as shown in equation \ref{eq:2}, resulting in a metric that accurately measures both the segmentation quality and the safety-related considerations of driving scene applications.

    Let $C$ be the set of all classes at the bottom level of the hierarchy,  $d(c,s)$ be the tree distance between class $c$, $s$, and $n$ be the number of levels in the hierarchy, $\text{gt}_c$ the set of pixels in the ground truth with label $c$ and $\text{pred}_s$ the set of pixels in the prediction with label $s$. We define the following quantities:
\begin{align}
I_{c,s}^{\text{safe}} &= \frac{ | \text{gt}_c \cap \text{pred}_s |}{ | \text{gt}_c \cup \text{pred}_c |} \tag{a} \qquad
I_{c,c} &= \frac{ | \text{gt}_c \cap \text{pred}_c |}{ | \text{gt}_c \cup \text{pred}_c |} 
\end{align}

Now, we define SmIoU as follows:
 
\begin{equation} \label{eq:1}
 I_c^{\text{safe}}  = 
\begin{cases}
\begin{aligned}[t]
        I_{c,c} & - \sum_{s \in C , s \neq c}\frac{d(c,s)}{n}I_{c,s}^{\text{safe}} & \text{ if } c \in C_{\text{imp}} \\
        I_{c,c} & - \sum_{s \in C_{\text{imp}} }\frac{d(c,s)}{n}I_{c,s}^{\text{safe}}& \text{ else. }
\end{aligned}
\end{cases} 
\end{equation}

\begin{equation}\label{eq:2}
\text{SmIoU} = \frac{\sum_{c\in C} I_c^{\text{safe}}}{|C|}
\end{equation}

Note that the definition of SmIoU requires the definition of important classes $C_{\text{imp}}$. When SmIoU is mentioned without specifying $C_{\text{imp}}$, we take it to be the union of traffic participants and roadside object classes (all classes except Far Objects and Sky in Figure \ref{fig:idd_ds_pixel_comparison}).

\begin{table}[t]
\centering
\resizebox{\columnwidth}{!}{
\begin{tabular}{@{}lllllllll@{}}\toprule

     \shortstack{Dataset\\[0.3em] Test $\rightarrow$ \\[0.3em] Train $\downarrow$} & \rot{CS} & \rot{ACDC}  &\rot{IDD}& \rot{Rain} & \rot{Fog} & \rot{LL } & \rot{Snow} & \rot{IDD-AW}  \\
     
    \midrule
     CS RGB& 83 & - & - & 46 &  45 & 42 & 43 & \cellcolor{gray!20}46\\
     ACDC RGB& - & 75 & - & 47 & 51 & 42 & 38 & \cellcolor{gray!20}48\\
     IDD RGB&- & - & 73 & 52 & 55 & 50 & 33 & \cellcolor{gray!40}54\\

     IDD-AW RGB &\cellcolor{gray!20}49 &\cellcolor{gray!20}51 &\cellcolor{gray!40}51 & 62 & 64 & 62 & 53 & 64 \\
    
     IDD-AW NIR & - & - & - & 61 & 58 & 57 & 51 & 61 \\
    
     IDD-AW NIR+RGB & - & - & - &\textbf{66} & \textbf{65} & \textbf{63} & \textbf{53} & \textbf{67}\\

    \bottomrule

  \end{tabular}
}
\caption{Comparison of mIoU scores (\%) of InternImage-b model trained on
CityScapes (CS), ACDC, IDD, and IDD-AW  datasets (RGB, NIR, and Combined) for IDD-AW  test set in individual conditions and jointly for all conditions. The model trained on IDD-AW  NIR+RGB gives the highest accuracy which is 20\% more than CS, ACDC, and 14\% more than IDD. The NIR+RGB model gives 3\% more accuracy compared to the RGB model, which indicates that the NIR image adds useful information for prediction. Also, note that the model trained on IDD-AW gives 3\%  better accuracy when tested on CS and ACDC as compared to the models trained on the respective datasets tested on IDD-AW (in grey). This trend is reversed in IDD since its training set is double the size of IDD-AW (in darker grey).
}
\label{tab:miou-comparison}
\end{table}

\paragraph{Properties of SmIoU.} 
The following are the important properties of the proposed SmIoU when compared with traditional mIoU, which show their relationship and motivate the definition of SmIoU for capturing unsafe mispredictions.
$ $\\

\begin{enumerate}
    \item \textit{$\text{SmIoU} = \text{mIoU}$, when $C_{\text{imp}} = \emptyset$.} 
When there are no important classes, SmIoU matches traditional mIoU. This is because the penalty becomes zero from the above formula, resulting in SmIoU becoming equal to the original mIoU.
\item \textit{$\text{SmIoU} \in [-1,1]$.} As we are penalizing the mIoU with the tree distance divided by the number of levels in the hierarchy, it always ranges between -1 to 1 inclusive of both extremes.
\item  \textit{$\text{SmIoU} = 1$} if and only if all pixels in an image are accurately classified into their respective classes.
\item \textit{$\text{SmIoU} = -1$} if and only if all pixels in an image are misclassified with the farthest classes in the tree hierarchy.
\end{enumerate}

\begin{table}[t]
\centering
\begin{tabular}{@{}lcccccccc@{}}\toprule

eval $\rightarrow$&\multicolumn{4}{c}{cross}&&\multicolumn{3}{c}{same}
\\\cmidrule{2-5} \cmidrule{7-9}
    \shortstack{Test $\rightarrow$ \\[0.3em] Train $\downarrow$} & \rot{Rain} & \rot{Fog} & \rot{LL} & \rot{Snow} & &\rot{mIoU}&\rot{\shortstack{SmIoU\\(tp)}} & \rot{SmIoU}\\
    \midrule
    IDD  &  52 & 55 & 50 & 33&&-&-&- \\
    IDD-AW & - & - & - & - && 64&60&51  \\
    \cmidrule{1-9}
    Rain & - & 55 & 40 & 29&&64&58&48\\
    Fog & 51 & -  & 53 &  29&&64&58&47 \\
    LL & 52 & 57 & - & 30&&62& 58&48  \\
    Snow & 35 & 38 & 33 & -&&53& 43 &28\\
    \bottomrule
\end{tabular}
\caption{Cross-evaluation of InternImage-b condition experts on
the various conditions of IDD-AW are shown in the first 4 columns. Comparison of mIoU (\%) with SmIoU (\%) metric at different levels and label sets for various adverse weather conditions shown in the last 3 columns. Here, SmIoU refers to Safe mIoU and tp refers to setting the important classes to be just the traffic participants (ie Living Things $\cup$ Vehicles in Figure \ref{fig:idd_ds_pixel_comparison}).}
\label{cond-miou}
\end{table}

\section{Results}
\label{sec:results}

\paragraph{Training on CS, IDD, and ACDC.} We have performed several experiments on Cityscapes, ACDC, IDD, and IDD-AW  datasets using the InternImage \cite{internimage} model. InternImage was chosen due to its state-of-the-art performance currently in semantic segmentation benchmarks. We have taken InternImage-b backbones of the InternImage \cite{internimage}. Regarding the results from Table \ref{tab:miou-comparison}, we can see the mIoU when trained on Cityscapes does not even cross 45\% mIoU when tested on IDD-AW. As the underlying dataset does not have any images with different weather conditions in it, the performance of a CS-trained model on IDD-AW  is very low. Similarly, a model trained on ACDC does not cross 50\%, which shows the dataset is that much harder and has a larger label set. Even though ACDC is also a driving dataset that is captured in multiple weather conditions, there's still not much improvement from Cityscapes. This already shows that our dataset is much harder than both Cityscapes and ACDC datasets.

When using pre-trained models of the IDD dataset, it performs much better than Cityscapes and ACDC despite testing with a larger label set. This is due to the unstructured images being present in the IDD dataset. The IDD pre-trained models performed efficiently in well-lit conditions like rain and fog, nonetheless the performance did not cross the 50\% mIoU mark in low-light conditions. Also, the major concern is the performance in snow conditions, where the best IDD pre-trained model, gives only 32\% mIoU which is a significant drop and is not usable for this condition at all. So even though the final mIoU reaches around 53\%, it cannot be generalized over each condition and overly fails to perform in certain conditions.

\begin{figure}[t]
\centering
\includegraphics[width=\columnwidth]
{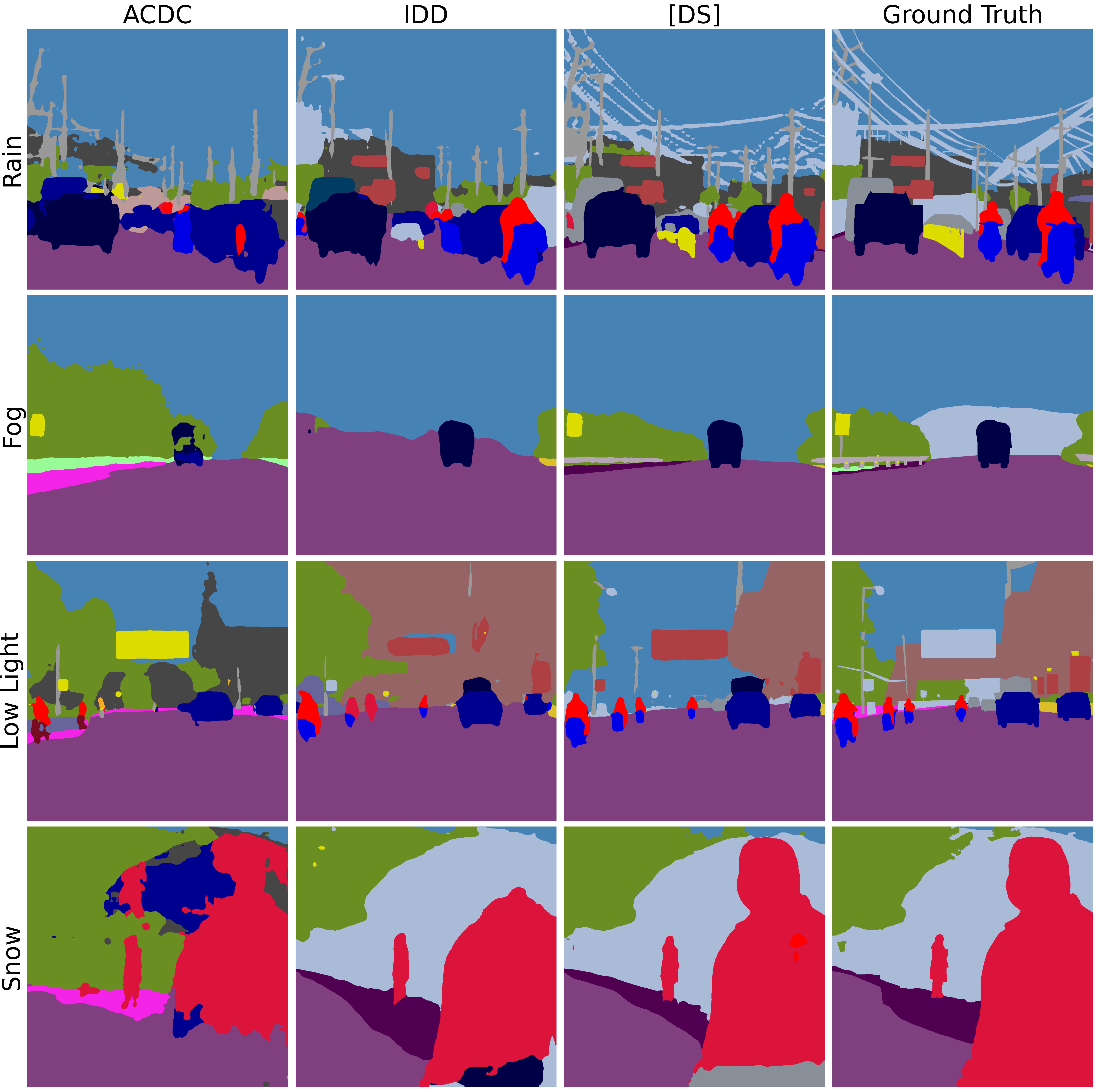} 
\caption{Qualitative examples from each condition with predictions using models trained on ACDC, IDD, and IDD-AW  Datasets and the ground truth at the end. The mIoUs (\%) for each pre-trained model on the IDD-AW  test set are 46.57, 53.4, and 64.5 using ACDC, IDD, and IDD-AW pre-trained models respectively.}

\label{fig:preds_comp}
\end{figure}

\paragraph{Training and Evaluation on IDD-AW.}
We perform the training in three different ways on IDD-AW. First, we train the network only on the RGB images of our dataset. We can observe that training on IDD-AW  gives more than 10\% mIoU score when compared to IDD and almost 20\% when taking the best performing Cityscapes or the ACDC pre-trained model. This is because of the variety of images in our dataset in various weather conditions which are not present in regular driving datasets like IDD or Cityscapes, as well as the unstructured nature of images and varying traffic when compared to other adverse datasets like ACDC. From Figure \ref{fig:preds_comp}, we can see qualitative results from each pre-trained model on our dataset.

We conducted a similar experiment using only the NIR components of the images, yielding a mIoU of over 60\% across various models and highlighting the quality of both the RGB and NIR components. Furthermore, we include training with stacked RGB+NIR images as input. Notably, the combination of RGB and NIR images through stacking led to a significant increase in mIoU by more than 3\% when compared to using just RGB or NIR components alone (refer to Table \ref{tab:miou-comparison}), underscoring the critical role of NIR components within our dataset.

\paragraph{Condition Expert Training.}
In condition expert training, we train specifically on each condition separately and then test on each condition to assess the quality of images across each set. In Table \ref{cond-miou}, we can see the differences between training on one condition and testing on each of them. We can see that training on any other condition except Snow, does not yield any significant mIoU increase in the snow images.

\paragraph{Safe mIoU on IDD-AW  test set.}

\begin{figure}[t]
\centering
\includegraphics[width=\columnwidth]{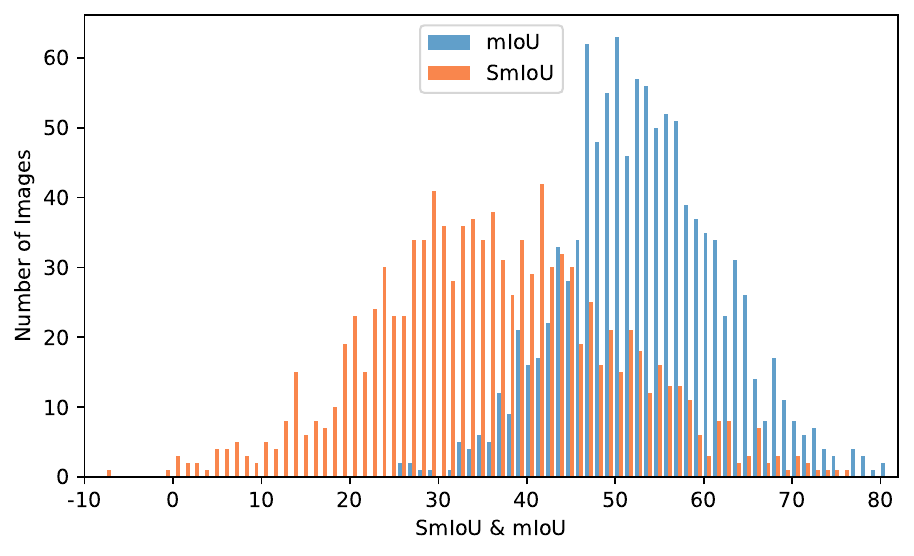} 
\caption{Distribution of Safe mIoU (\%) vs mIoU (\%) for IDD-AW  test set. y-axis represents the number of images with a particular value of SmIoU/mIoU. The SmIoU distribution is shifted to the lower side, indicating that it is finding a significant amount of dangerous mispredictions that are not accounted for in mIoU.}
\label{fig:smiou_miou_hist}
\end{figure}

\begin{table*}[t]
\centering
\resizebox{\textwidth}{!}{
\begin{tabular}{@{}llrcrrrrrrrrrrrrrrrr@{}} \toprule
\rot{Metric}
&
    \rot{Condition} 
    & \rotatebox{90}{road}
    &\rot{\begin{tabular}[c]{@{}l@{}}drivable\\ fallback\end{tabular} }
    &\rotatebox{90}{sidewalk}
    &\rotatebox{90}{person}
    &\rotatebox{90}{rider}
    &\rotatebox{90}{bike} 
    &\rotatebox{90}{bicycle}
    &\rotatebox{90}{rickshaw}
    &\rotatebox{90}{car}
    &\rotatebox{90}{truck}
    &\rotatebox{90}{bus}
    &\rotatebox{90}{\begin{tabular}[c]{@{}l@{}}vehicle\\ fallback\end{tabular}}
    &\rotatebox{90}{curb}
    &\rotatebox{90}{wall}
    &\rotatebox{90}{guard rail}
    &\rotatebox{90}{billboard}
    &\rotatebox{90}{\begin{tabular}[c]{@{}l@{}}traffic\\ sign\end{tabular}}
    &\rotatebox{90}{\begin{tabular}[c]{@{}l@{}}traffic\\ light\end{tabular}} \\
    \midrule
    \multirow{5}{*}{\rot{mIoU \%}}&
    All & 95 & 51 & 48 & 76 & 72 & 68 & 5 & 83 & 85 & 74 & 76 & 45 & 78 & 52 & 74 & 62 & 58 & 52 \\
    &Rain & 96 & 49 & 49 & 48 & 73 & 68 & 2 & 86 & 87 & 79 & 52 & 47 & 81 & 47 & 46 & 64 & 54 & 52 \\
    &Fog & 97 & 64 & 24 & 62 & 75 & 73 & 8 & 67 & 91 & 82 & 80 & 49 & 79 & 57 & 78 & 61 & 69 & 47 \\
    &LL & 95 & 60 & 54 & 73 & 73 & 68 & 38 & 75 & 80 & 69 & 86 & 29 & 65 & 50 & 72 & 58 & 35 & 53 \\
    &Snow & 85 & 42 & - & 80 & 62 & 40 & 0 & - & 82 & 58 & 70 & 48 & 23 & 56 & 64 & 60 & 37 & -\\
    \midrule
    \multirow{5}{*}{\rot{SmIoU \%}} &
    All & 92 & 32 & 16 & 64 & 58 & 52 & - 22 & 77 & 81 & 68 & 70 & 21 & 69 & 32 & 61 & 42 & 40 & 27 \\
    
    &Rain & 94 &  28 &  25 &  15 &  59 &  54 &  -7 &  81 &  84 &  74 &  46 &  23 &  74 &  22 &  20 &  46 &  32 &  26  \\
    
    &Fog & 95 &  51 &  -48 &  41 &  62 &  58 &  -63 &  43 &  87 &  78 &  77 &  28 &  71 &  52 &  69 &  45 &  53 &  12 \\
    
    &LL & 92 & 46 & 22 & 60 & 59 & 52 & 14 & 70 & 76 & 61 & 82 & 2 & 48 & 26 & 52 & 35 & 15 & 32\\
    
    &Snow & 79 & 20 & - & 70 & 44 & -2 & -99 & - & 76 & 48 & 56 & 23 & -46 & 23 & 42 & 38 & 5 & -  \\
    \bottomrule
  \end{tabular}
}
\caption{Comparison of class-wise labels for important classes between mIoU vs SmIoU for InternImage-b model on segmentation.}

\label{miou_vs_safemiou_classwise}
\end{table*}

 In Table {\ref{cond-miou}}, we can see the drop in SmIoU when we consider the important classes. For the InternImage-b framework, the SmIoU again drops more than 15\% at least for each individual condition. This clearly shows the discrepancy and the dangers of traditional mIoU in hierarchical autonomous driving datasets. In Figure \ref{fig:smiou_miou_hist}, we can see the distribution of SmIoU vs mIoU over the test set has shifted toward the lower side which some images exhibiting even negative SmIoU. 

 Notice that SmIoU = mIoU if $C_{\text{imp}} = \emptyset$. When we add more classes to $C_{\text{imp}}$ the metric value decreases. This can be observed in the last 3 columns of Table \ref{cond-miou}. They correspond to $C_{\text{imp}} = \emptyset$, traffic participants and traffic participants $\cup$ roadside objects respectively. The SmIoU (tp) has values between mIoU and SmIoU. This shows our new metric is flexible and very effective at identifying the images in which the important classes are misclassified.

When comparing class-wise mIoUs and SmIoUs for the InternImage-b framework as in Table {\ref{miou_vs_safemiou_classwise}}, we can see several classes showing great disparities between mIoU and SmIoU. There is a significant difference between things like bicycles, traffic signals, and sidewalks. It can be dangerous to misclassify these classes, especially when driving. When compared to conventional mIoU, our suggested SmIoU metric more accurately measures this risk. SmIoU also shows several classes with negative values showing dangerous misclassification, especially bicycle, sidewalk, and curb. The danger level associated with misclassification that could be brought on by these objects in the drive scenes is represented by classes like person in rain, sidewalk, and rickshaw in fog, vehicle fallback in lowlight, bike, and curb in snow that is well captured by SmIoU but missed by the traditional mIoU.


\section{Conclusion}

We have presented IDD-AW, a large-scale
dataset and a benchmark suite for semantic driving scene understanding in adverse 
 weather and unstructured driving conditions. We also present a new metric called Safe mIoU which incorporates safety concerns in the definition of mIoU.  We benchmark state-of-the-art models for semantic segmentation in IDD-AW  and also show the differences between traditional mIoU and safe mIoU while considering important classes. Finding appropriate loss functions, which can better optimize safe mIoU more efficiently is an interesting direction for future work.

\section{Acknowledgement}
We acknowledge the funding from DYSL-AI, DRDO for this project. The dataset was collected in collaboration with \href{https://mobility.iiit.ac.in/}{iHub Data}, IIIT Hyderabd.

{\small
\bibliographystyle{ieee_fullname}
\bibliography{ref}
}
\pagebreak

\end{document}